\documentclass[conference]{IEEEtran}
\IEEEoverridecommandlockouts
\usepackage{cite}
\usepackage{amsmath,amssymb,amsfonts}
\usepackage{algorithmic}
\usepackage{graphicx}
\usepackage{textcomp}
\usepackage{xcolor}
\usepackage{tikz}
\usepackage{url}
\usepackage{subcaption}
\usepackage{hyperref}

\hypersetup{
  bookmarks=true,     
  bookmarksopen=true, 
}
\def\BibTeX{{\rm B\kern-.05em{\sc i\kern-.025em b}\kern-.08em
    T\kern-.1667em\lower.7ex\hbox{E}\kern-.125emX}}
\begin{document}

\title{Sparse Autoencoder Insights on Voice Embeddings}

\author{\IEEEauthorblockN{Daniel Pluth}
\IEEEauthorblockA{\textit{Vail Systems, Inc.}\\
Chicago, USA \\
dpluth@vailsys.com}
\and
\IEEEauthorblockN{Yu Zhou}
\IEEEauthorblockA{\textit{Vail Systems, Inc.}\\
Chicago, USA \\
yzhou@vailsys.com}
\and
\IEEEauthorblockN{Vijay K. Gurbani}
\IEEEauthorblockA{\textit{Vail Systems, Inc.}\\
Chicago, USA \\
vgurbani@vailsys.com}

}

\maketitle

\begin{abstract}
Recent advances in explainable machine learning have highlighted the potential of sparse autoencoders in uncovering mono-semantic features in densely encoded embeddings. While most research has focused on Large Language Model (LLM) embeddings, the applicability of this technique to other domains remains largely unexplored. This study applies sparse autoencoders to speaker embeddings generated from a Titanet model, demonstrating the effectiveness of this technique in extracting mono-semantic features from non-textual embedded data. The results show that the extracted features exhibit characteristics similar to those found in LLM embeddings, including feature splitting and steering. The analysis reveals that the autoencoder can identify and manipulate features such as language and music, which are not evident in the original embedding. The findings suggest that sparse autoencoders can be a valuable tool for understanding and interpreting embedded data in many domains, including audio-based speaker recognition.
\end{abstract}


\section{Introduction} \label{sec:intro}

Sparse autoencoders (SAEs) have been used to uncover features in densely encoded
vectors to great success. Most research in this area has focused on
embeddings generated from large language models, showing good mono-semantic
feature extraction. However, many complex neural network models will generate
embedded data prior to delivering the desired target. In principle, the same
SAE technique could be applied to many different forms of data.
Explainable machine learning models are critical to understand what the model is
acting on to make predictions. When minor perturbations can result in different
predictions, that indicates that deep neural networks may be acting on
undesirable inputs\cite{Su_2019}. Insight into the embedded data created by
these models would help alleviate some of their black-box nature, and allow a
better understanding of how features are being interpreted by a given model.

In the present work, a speaker embedding model is used to
encode audios into embeddings which represent speaker characteristics. These
embeddings are examined through the latent space of an SAE. The elements of the latent space are discovered to be mono-semantic, meaning that their activation corresponds to a singular meaningful characteristic of the original audios. 

The main contribution of this work is showing that SAEs can be used
effectively for non-transformer based models, that the method can be used to
extract mono-semantic features from audio-based speaker data, and
that the features demonstrate the same characteristics discovered in LLM-based
studies, such as feature splitting and feature steering.


\section{Related Work}
\label{sec:related-work}
There has been much work to improve the interpretability of neural network
models in the past
\cite{ribeiro2016whyitrustyou,
        zafar2019dlimedeterministiclocalinterpretable,
        shrikumar2019learningimportantfeaturespropagating,
        sundararajan2017axiomaticattributiondeepnetworks}.
    LIME-based solutions look at the output of the models to better
    understand what they're acting upon, but do not investigate how those decisions
    are being made. Other solutions that examine model weights work quite well
    for image recognition and other tasks where reconstructed input data can be estimated
    or otherwise interpreted, however few of these have been put to use for
    audio data.
    
    One issue with audio may be the very nature of the data, it can be difficult to visualize and understand small variations. There have been
    attempts to improve interpretability for speaker recognition models. One
    straightforward solution is to directly use the speaker embedding generated
    by a model to perform additional classification
    tasks\cite{luu2021leveragingspeakerattributeinformation}. This method
    demonstrates that the model has sufficient information to make various
    classifications, but not that the information is necessarily impactful to
    the speaker recognition task. Another method trained similar classification
    tasks as a first stage, and then the speaker verification task was trained
    on the output of the first stage\cite{wu2024explainableattributebasedspeakerverification}. The issue here being that the model is
    then constrained to only act on the attributes manually selected by the
    researcher.
    
    Recently, several studies have utilized SAEs as a method
    to examine the data embedded within the tokens utilized in Large
    Language Models (LLMs) \cite{bricken2022monosemanticity, templeton2024scaling, gao2024scalingevaluatingsparseautoencoders, lieberum2024gemmascopeopensparse}. Since the method acts directly on the embedded data, any given hidden layer output could be examined. This is a promising new technique that may allow better examination of densely encoded data.

\section{Method}
\label{sec:method}
The purpose of this study is to investigate if a sparse autoencoder model is capable of extracting mono-semantic features from non-textual embedded data.
Speaker biometric data is chosen as it is quite distinct from LLM embeddings.
This type of embedding is interesting because the data it represents is more
continuous and the embeddings are generated in a entirely different manner
than those of an LLM.

In order to investigate these mono-semantic features, several steps are
required. First, a large set of speaker characteristic embeddings are generated.
These embeddings are then used to train a series of SAEs. Using
these autoencoders, the speaker embeddings are transformed into sparse latent
spaces. The characteristics and behaviors of these latent spaces are then
investigated.

\subsection{Speaker Embedding}
\label{sec:embedding}
The embeddings are generated using a fine-tuned version of the
Titanet model\cite{koluguri2021titanetneuralmodelspeaker}. Titanet uses convolution
layers and statistical pooling in order to transform utterances of variable
length into a 192 element vector which can be used as a representation of the
speaker's characteristics. These vectors, also known as embeddings, are dense and not clearly interpretable. They are suitable for speaker identification and recognition, but characteristics like language, pitch, and gender are non-obvious in the embeddings. For this work, the NeMo framework\cite{Harper_NeMo_a_toolkit} is used to fine-tune Titanet to create a telephony-adapted variant. The resultant model is evaluated on a withheld set of speakers
and achieved a sub 1\% equal error rate (EER) in the telephony domain.
This is the model used to
generate speaker embeddings to train the autoencoders described below.

\subsection{Sparse Autoencoder}
\label{sec:sae}
A SAE is used to disentangle features from Titanet
embeddings. An autoencoder typically transforms an input feature set into a
latent space and then back into a reconstruction of the original feature. During training, the mean squared error can be used to drive reconstruction accuracy and ensure that the autoencoder has learned the relationships within the data.

\begin{figure}
    \centering
    \usetikzlibrary{fit}
    \usetikzlibrary{shapes.geometric}
    \usetikzlibrary{positioning}
\begin{tikzpicture}[
    node distance = 2cm,
    box/.style = {rectangle, draw, minimum width=1cm, minimum height=1cm},
    latent/.style = {rectangle, draw, fill=red!30, minimum width=1cm, minimum height=3cm},
    arrow/.style = {->,>=stealth,thick},
    label/.style = {text width=3cm, align=center, font=\small},
]
\node[box, fill=gray!20] (input) at (0,0) {$\mathbf{e}$};
\node[box, fill=blue!10] (output) at (6,0) {$\mathbf{\epsilon}$};

\node[trapezium, rotate=90, fill=green!20] (encoder) at (2,0) {Encoder};
\node[trapezium, rotate=270, fill=blue!20] (decoder) at (4,0) {Decoder};

\node[latent] (latent) at (3,0) {$\mathbf{v}$};
\draw[arrow] (input) -- (encoder);
\draw[arrow] (encoder) -- (latent);
\draw[arrow] (latent) -- (decoder);
\draw[arrow] (decoder) -- (output);

\node[draw, dashed, fit=(encoder) (latent) (decoder), label=above:Autoencoder] (ae) {};
\node[label, above=1.6cm] at (ae) {Sparse Autoencoder};

\end{tikzpicture}
    \caption{Embedding $e$ is reconstructed as $\epsilon$ via latent vector $v$.}
    \label{fig:arch}
\end{figure}
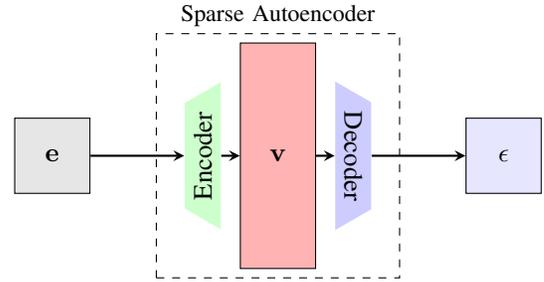

While the latent space in a typical autoencoder is smaller than the original feature space, for an SAE as used in this work, the inverse is performed in order to
explode the dimensionality of the latent space to many times the original feature size. The general layout used in the present work is shown in Figure \ref{fig:arch}. It is important to note that without additional constraints a naive autoencoder with a latent dimension greater than or equal to the feature space could simply act as a
pass-through for the feature data. In order to enforce sparsity and reduce any pass-through effect, two methods have
previously been utilized in the literature: TopK activation or L1 regularization
\cite{gao2024scalingevaluatingsparseautoencoders, templeton2024scaling}. L1
regularization alongside a ReLU activation function helps to reduce the total
activations of the latent layer and lead to dropping non-critical latent elements during the course of the training. However, this method can have a
significant issue with 'dead latents' or elements of the latent layer that
never activate once the model is trained. The use of a TopK activation after
the encoding layer can reduce the dead latents effect significantly, however
brings about an added complication of an additional parameter that needs to be
tuned. There does not appear to be a good method to calculate the ideal values
for latent dimension and TopK value. In this study a grid search is performed
and many models with different combinations of activations and latent dimension are trained. All models examined are well trained, with a stable validation mean squared error.


\subsection{Feature Identification}
The latent space of the  SAE can in principle be examined directly
by examining samples which share a common latent activation and then listening
to them to identify shared characteristics. This proved to be far more difficult
than anticipated as most of the latent elements correspond to features that are completely non-obvious in the audio itself. The challenges of this approach are detailed in Section
\ref{sec:discussion}.

Instead, a dataset is curated with the desired feature labeled
manually and then a logistic regression model was trained on the latents in order to classify the feature, as described in \cite{templeton2024scaling}. The
weights of the logistic regression model are then used to identify which of the
latents is most important for making the classification. The latent can then be used as a
predictor for the desired feature and its performance can be evaluated on a withheld test
set.

Two primary attributes are examined using this method, the language of the
speaker and music-only audio. These are chosen due to their ease of unambiguous
labeling. Music is present in this dataset as hold music which is often played during
a given phone call. This music is never coincident with the caller's voice.

\subsection{Feature Steering}
\label{sec:steer}
In addition to measuring the discriminant accuracy, the effect of an identified latent feature can also be demonstrated with feature steering, where the activation of the feature is artificially turned on or off to observe how the reconstructed speaker embedding is impacted.

For example, one can measure the similarities of a given speaker embedding to the reference Spanish and English embeddings before and after the activation of the latent feature identified as Spanish audio is manipulated.  Formally, given an SAE model (Fig.\ref{fig:arch}) that encodes an input speaker embedding $\mathbf{e} \in \mathbb{R}^M$ into a latent vector $\mathbf{v} \in \mathbb{R}^L$, and subsequently decodes $\mathbf{v}$ into a reconstructed speaker embedding $\mathbf{\epsilon} \in \mathbb{R}^M$, 
if the latent index $\phi \in [0,L-1]$ of the SAE model has been identified as the feature that signifies Spanish audio samples when activated, then the significance of this feature can be examined by deactivating it for the latent vector of a Spanish sample $\mathbf{v}^{(S)} \in \mathbb{R}^L$ in the test set, with $\mathbf{v}^{(S)}_\ell$ denoting element $\ell \in [0,L-1]$ of $\mathbf{v}^{(S)}$:
\begin{equation}
\tilde{\mathbf{v}}^{(S)}_\ell = 
\begin{cases}
\mathbf{v}^{(S)}_\ell  & \text{if } \ell \ne \phi \\
-a_\phi   & \text{if } \ell = \phi 
\end{cases}
\label{eq:steer_spanish}
\end{equation}

Conversely, feature $\phi$ can be artificially activated for the latent vector of an English sample in the test set $\mathbf{v}^{(E)}$:
\begin{equation}
\tilde{\mathbf{v}}^{(E)}_\ell = 
\begin{cases}
\mathbf{v}^{(E)}_\ell  & \text{if } \ell \ne \phi \\
a_\phi   & \text{if } \ell = \phi 
\end{cases}
\label{eq:steer_english}
\end{equation}

where $a_\phi \in \mathbb{R}^{+}$ in Eqs.(\ref{eq:steer_spanish}) and (\ref{eq:steer_english}) is a pre-selected value for manually steering the latent feature $\phi$.  In this study, we choose $a_\phi=1$.


Finally, the feature-steered speaker embedding $\tilde{\mathbf{\epsilon}}$ can be constructed by the SAE decoder using $\tilde{\mathbf{v}}$ from Eq.(\ref{eq:steer_spanish}) or (\ref{eq:steer_english}) as input.  In order to evaluate the difference between pre-steering $\epsilon$ and post-steering  $\tilde{\mathbf{\epsilon}}$, their relative similarity scores are computed and compared:
\begin{equation}
\delta_s(\mathbf{x}) = s(\mathbf{x},\hat{\mathbf{\epsilon}}^{(S)}) - s(\mathbf{x},\hat{\mathbf{\epsilon}}^{(E)}) 
\label{eq:relative_sim}
\end{equation}

where $\mathbf{x} \in \{\epsilon, \tilde{\mathbf{\epsilon}}\}$. 
 In Eq.(\ref{eq:relative_sim}), $\hat{\mathbf{\epsilon}}^{(S)}$ and $\hat{\mathbf{\epsilon}}^{(E)}$ are the centroids of SAE-reconstructed Spanish and English embeddings in the training set, respectively, and $s(\mathbf{x},\hat{\mathbf{\epsilon}}^{(S)})$ denotes the cosine similarity score between $\mathbf{x}$ and $\hat{\mathbf{\epsilon}}^{(S)}$.  
 Therefore the relative similarity score $\delta_s(\mathbf{x})$ measures how much closer a speaker embedding $\mathbf{x}$ is to 
the reference Spanish embedding $\hat{\mathbf{\epsilon}}^{(S)}$ than to the reference
English embedding $\hat{\mathbf{\epsilon}}^{(E)}$.  A positive $\delta_s(\mathbf{x})$
indicates a greater similarity to Spanish, while a negative value suggests
greater similarity to English.  The comparison of pre-steering $\delta_s(\mathbf{\epsilon})$ with post-steering $\delta_s(\tilde{\mathbf{\epsilon}})$ provides a quantitative view of the latent feature significance.

While the Spanish latent feature is used as an example in this section, Eqs.\ref{eq:steer_spanish} through \ref{eq:relative_sim} can be applied to all identified SAE latent features, with relative similarity score as a metric to measure the effectiveness of feature steering. 

\section{Experimental Setup}
\label{sec:setup}

\subsection{Dataset}
\label{sec:dataset}
The dataset used is a collection of phone calls to 
the customer support center. Music and automated announcements are occasionally captured as well, depending on the call. Due to the nature of these calls
the audios are private, however, the embeddings contain no private information.
The embeddings are created from 4-8 second long segments taken from the
original call using SoX to select non-silent sections of the audio. These audio
segments are processed via a custom trained telephony-based Titanet model to
generate speaker embeddings.

Three distinct datasets are curated for this experiment: 1) Autoencoder training dataset, 2) Language Feature dataset, 3) Music Feature dataset. The Autoencoder training dataset consists of approximately 1,100,000 Titanet embeddings from $\sim$29,000 speakers, each with a dimension of 192. This data has no test set as the autoencoders are trained in an unsupervised manner. The Language and Music Feature datasets are developed from a distinct set of audios which do not share the same speakers as the autoencoder training dataset.

The Whisper ASR model \cite{radford2023robust} is used to predict the spoken language, which functions as a loose label which is then manually verified, for the Language Feature
dataset. At the time of
annotation perceived gender was also tagged. The resultant demographics for this
dataset are shown in Table \ref{tab:language_gender_demographics}. 

\begin{table}[htbp]
    \centering
    \begin{tabular}{|c|c|c|}
    \hline
                       & Latent Element Training & Test \\ \hline
        English Female & 457   & 283 \\
        English Male   & 166   & 117  \\
        Spanish Female & 382   & 279  \\
        Spanish Male   & 166   & 121 \\
        \hline
    \end{tabular}
    \caption{Distribution of Language Feature dataset. The test set has an even balance of 400 samples per language.}
    \label{tab:language_gender_demographics}
\end{table}

The Music Feature dataset is labeled using transcripts generated from Whisper to
identify audios which have no voices. Since activity
detection is performed, these audios were not silent, rather they contain either noise or music. The resultant dataset is manually examined to identify music. Other spoken audios segments are used for the voice class. The distribution of this dataset is shown in Table
\ref{tab:music_voice_distribution}.

\begin{table}[htbp]
    \centering
    \begin{tabular}{|c|c|c|}
    \hline
         & Latent Element Training & Test \\ \hline
        Music & 109 & 200 \\
        Voice & 257 & 200 \\
        \hline
    \end{tabular}
    \caption{Distribution of training and test data for Music Feature dataset.}
    \label{tab:music_voice_distribution}
\end{table}

\subsection{Sparse Autoencoder}
A simple SAE model is developed based on the implementation of
OpenAI \footnote{\url{https://github.com/openai/sparse_autoencoder}}. The model
is a simple linear layer with an activation layer as the encoder to the latent
space. The decoder is another linear layer back to the feature space. Models are trained with both ReLU and TopK activations, however, ReLU models show a significant issue with dead latents, where much of the latent space was never active. Ultimately, only TopK activation is used. A grid search is performed with K value and latent dimension. Latent values between 100 and 1,200 are examined, however larger values are under-trained and omitted from the results. K values between 5 and 35 are also examined for TopK activation. 

\section{Results and Discussions} \label{sec:results}

\subsection{Feature Identification}
\label{sec:identify}

\subsubsection{Language} \label{sec:id_language} 
\begin{figure}[htbp] 
\centering
\begin{subfigure}{\columnwidth} \centering
\includegraphics[width=\columnwidth]{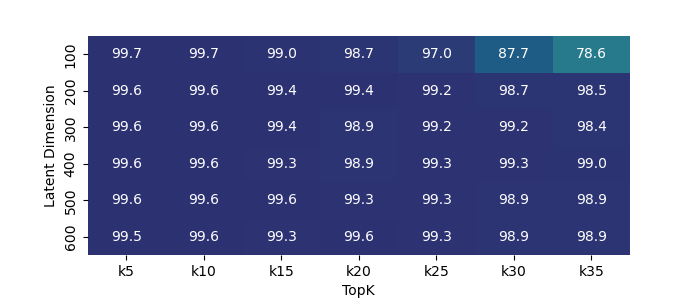}
\vspace{-5mm}
\caption{Precision} \end{subfigure} \\[1ex]
\begin{subfigure} {\columnwidth} \centering
\includegraphics[width=\columnwidth]{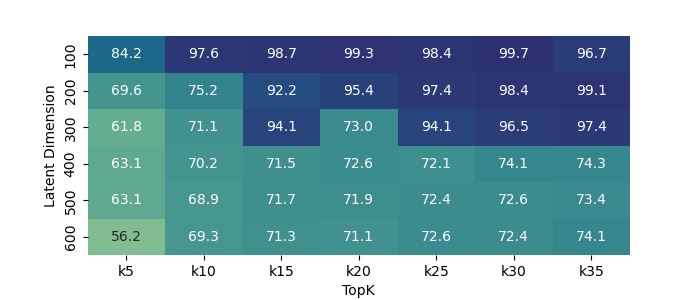}
\vspace{-5mm}
\caption{Recall}
\label{tab:spanish-index-performance-recall}

\end{subfigure}
\caption{Performance of the top latent index for classifying language across
models with varying latent dimension and TopK activation.}
\label{tab:spanish-index-performance}

\end{figure}

Using the training set of the Language Feature dataset and a logistic regression model, an index is identified for each given autoencoder model which most effectively discriminates between Spanish and English voices. 
A non-zero activation value for the given latent index is treated as Spanish. The capability of the index as a predictor is highly dependent on the model parameters as examined in the grid search. Figure \ref{tab:spanish-index-performance} shows the performance of the various models and the appropriate latent index in the task of Spanish
language identification. The majority of the models investigated have a recall that stabilizes around 70\%, however lower latent dimensions and higher k-values lead to recall values in the mid 90's. Further discussion of this result is found in Section \ref{sec:split}.

Taking as an example the model with K=$20$ and latent dimension of $200$, the latent index $15$ functions
as a discriminant with a precision of $99.2\%$ and recall of $95.5\%$.
Investigating the wrongly classified audios also leads to the discovery that
$91.0\%$ of the false positive
samples seems to be native Spanish speakers speaking English with the remainder being other non-native English speakers. The false negative set reveals the
occasional presence of code-mixing, in this case the use of English words in
Spanish speech.




\vspace{-5mm}



\subsubsection{IVR Music} \label{sec:id_music} 
\begin{figure}[htbp]
\centering
\subcaptionbox{Precision}{
\includegraphics[width=\columnwidth]{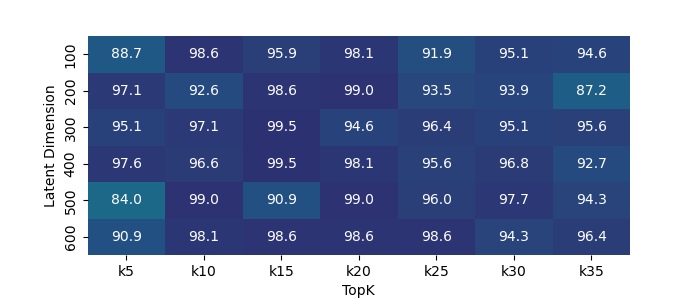} }
\subcaptionbox{Recall}{
\includegraphics[width=\columnwidth]{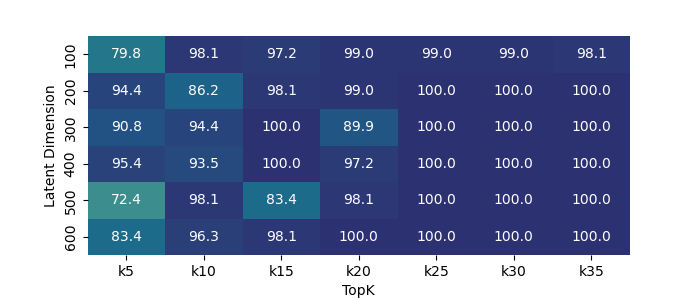} }
\caption{Performance of the top latent index for classifying music across
models with varying latent dimension and TopK activation.}
\label{fig:music-index-performance}
\end{figure}

Similar to Section \ref{sec:id_language}, the training set of the Music Feature dataset is used to identify
which element of the latent space is most significant in helping to identify
the desired class, in this case hold music. The leading latent activation works
as a high quality discriminator for music as seen in Figure
\ref{fig:music-index-performance}. Taking the same SAE model used in Section \ref{sec:id_language} with K=$20$ and latent dimension of $200$, the latent
index $74$ achieves a precision of $92.3\%$ and recall of $99.1\%$. It is
also clear that the performance is largely uniform for all SAEs that were examined. There is a drop in performance for models with a low K value. 



\subsection{Feature Splitting} \label{sec:split} 
\begin{figure*}[htbp] 
\centering
\includegraphics[width=0.8\textwidth]{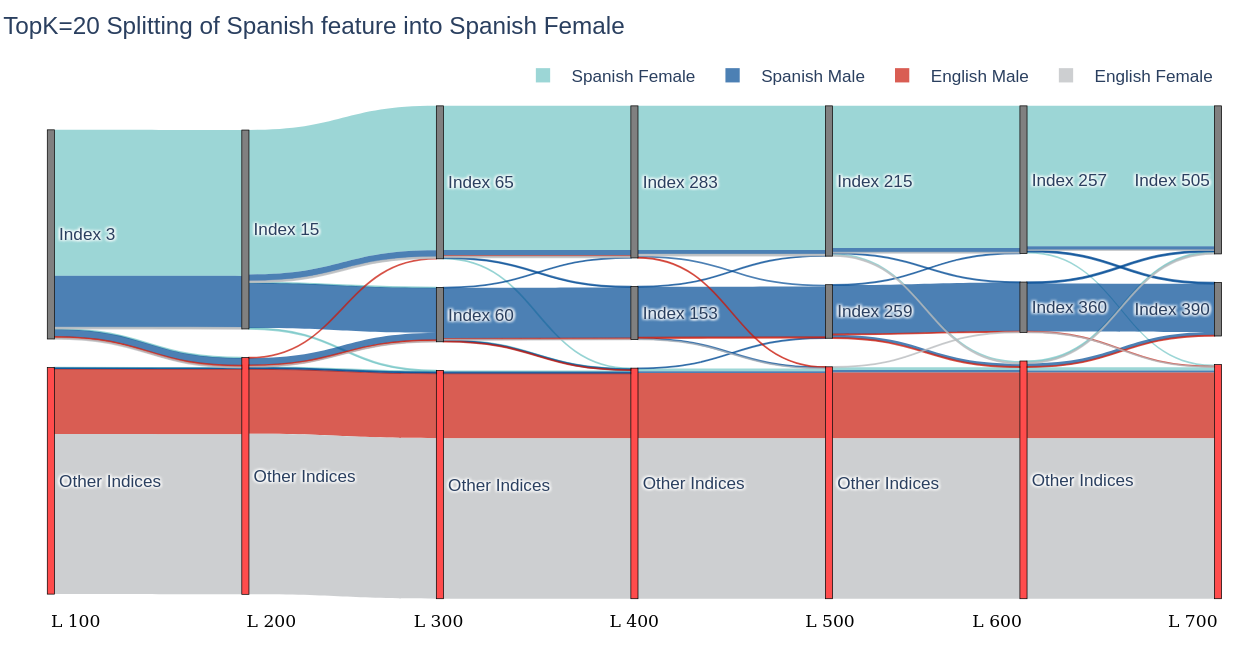}
\caption{The movement of the different language and gender samples in
and out of the predominant Spanish language index.}
\label{fig:spanish-sankey} 
\end{figure*}

Regarding Figure
\ref{tab:spanish-index-performance-recall}, it appears that there are two distinct
regimes of behavior. Further investigation reveals that these regimes are due
to a splitting of the Spanish language feature. When examining the audios
selected via the Spanish index in the lower recall model region, it becomes
apparent that the majority of the male Spanish speakers have dropped out of the
set. Figure \ref{fig:spanish-sankey} illustrates the splitting of the Spanish
feature as the latent space grows. For latents 100 and 200 the indices
3 and 15, respectively, are the Spanish language latent elements. However, for higher latent dimensionality, the Spanish language is no longer encoded as a single element of the latent space. Instead it is clear from the figure that the Spanish language feature splits into a Spanish male feature with index 60 and a Spanish female feature with index 65 at a latent of 300. This
same behavior repeats regardless of K value, however the latent dimensionality
at which it occurs does vary. Regarding Figure \ref{tab:spanish-index-performance-recall}, it appears that as K grows, the latent dimensionality required to cause a split also grows, but only up to a point.
After this point, which is around 300 elements, the behavior plateaus and the
split remains around the same latent dimensionality.

\subsection{Feature Steering} \label{sec:results_steer}
\subsubsection{Spanish vs. English} 
As discussed in Section \ref{sec:id_language}, in the SAE model with
latent dimension of 200 and $K=20$,  latent index 15 is identified as the
feature for Spanish, with attribution precision of 99.2\%.  
To examine its significance, feature steering as described in section \ref{sec:steer}
is performed by deactivating the latent feature for Spanish samples 
and activating it for English samples in the Language Feature test set.  

\begin{figure}[htbp] \centering
\includegraphics[width=\columnwidth]{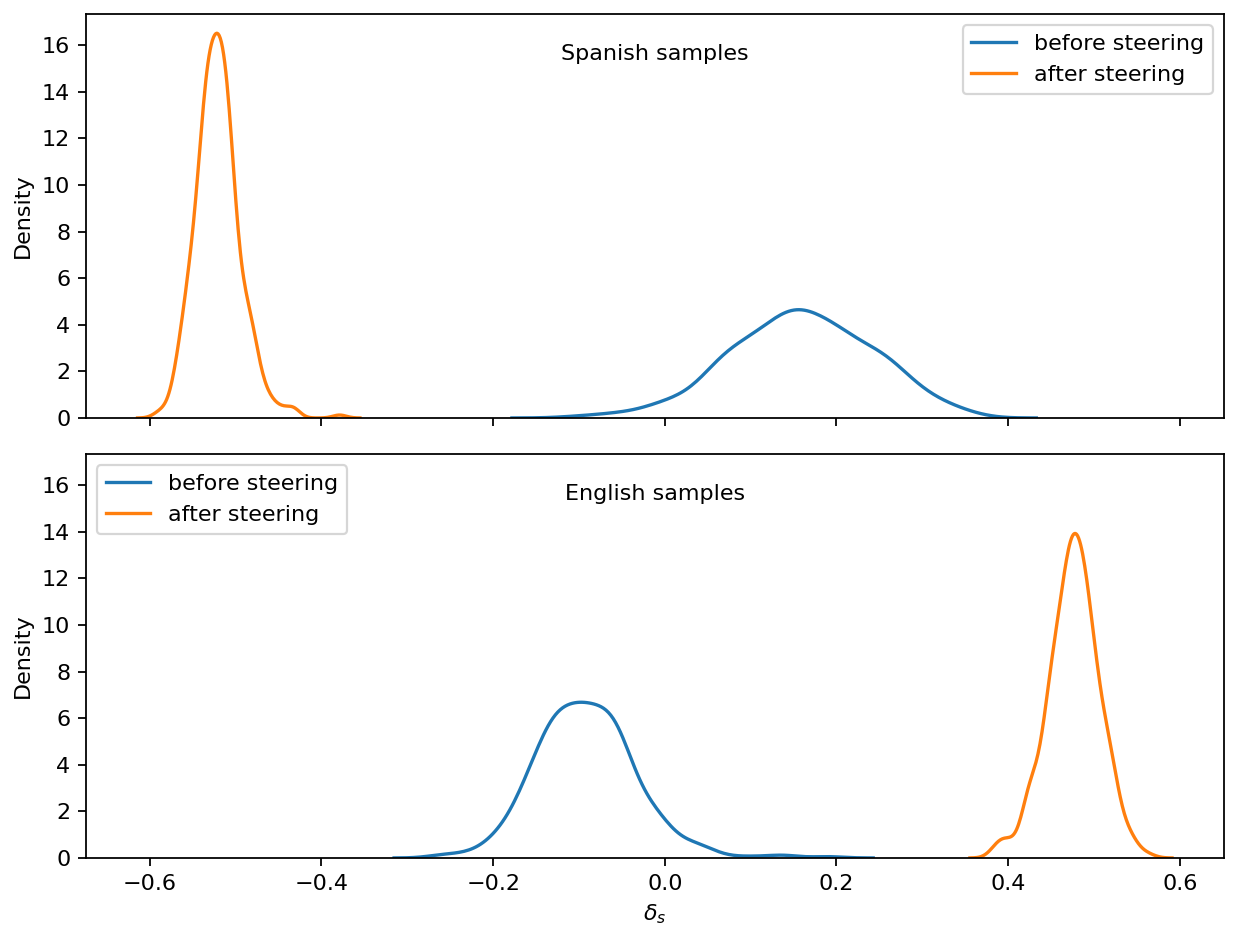}
\caption{Distribution of relative similarity scores before and
after Spanish feature steering.} 
\label{fig:language_steer} 
\end{figure}

Figure \ref{fig:language_steer} shows the distribution of relative similarity scores 
$\delta_s$ computed using Eq.(\ref{eq:relative_sim}) before and after feature steering 
for Spanish and English samples.  Since a positive $\delta_s$ signifies
the sample is more similar to the reference Spanish embedding than the reference English
embedding, it is observed that when this latent feature is 
deactivated for Spanish samples in the test set, the resulting SAE-reconstructed embeddings 
shift from the reference Spanish embedding $\hat{\mathbf{\epsilon}}^{(S)}$ toward the reference English embedding $\hat{\mathbf{\epsilon}}^{(E)}$.  
Conversely, when this latent feature is activated for English samples, the opposite 
shift occurs.  Table \ref{tab:language_steer} lists the mean values of the relative similarity
scores for Spanish and English samples before and after feature steering, note the change between positive and negative $\delta_s$ values.

\begin{table}[htbp]
    \centering
    \begin{tabular}{|c|c|c|}
    \hline
        & Before Steering & After Steering \\ \hline
        Spanish Samples & 0.160 & -0.520 \\
        English Samples & -0.091 & 0.475  \\
        \hline
    \end{tabular}
    \caption{Mean values of relative similarity scores for Spanish feature steering.}
    \label{tab:language_steer}
\end{table}

\subsubsection{IVR Music} 
\begin{figure}[htbp] \centering
\includegraphics[width=\columnwidth]{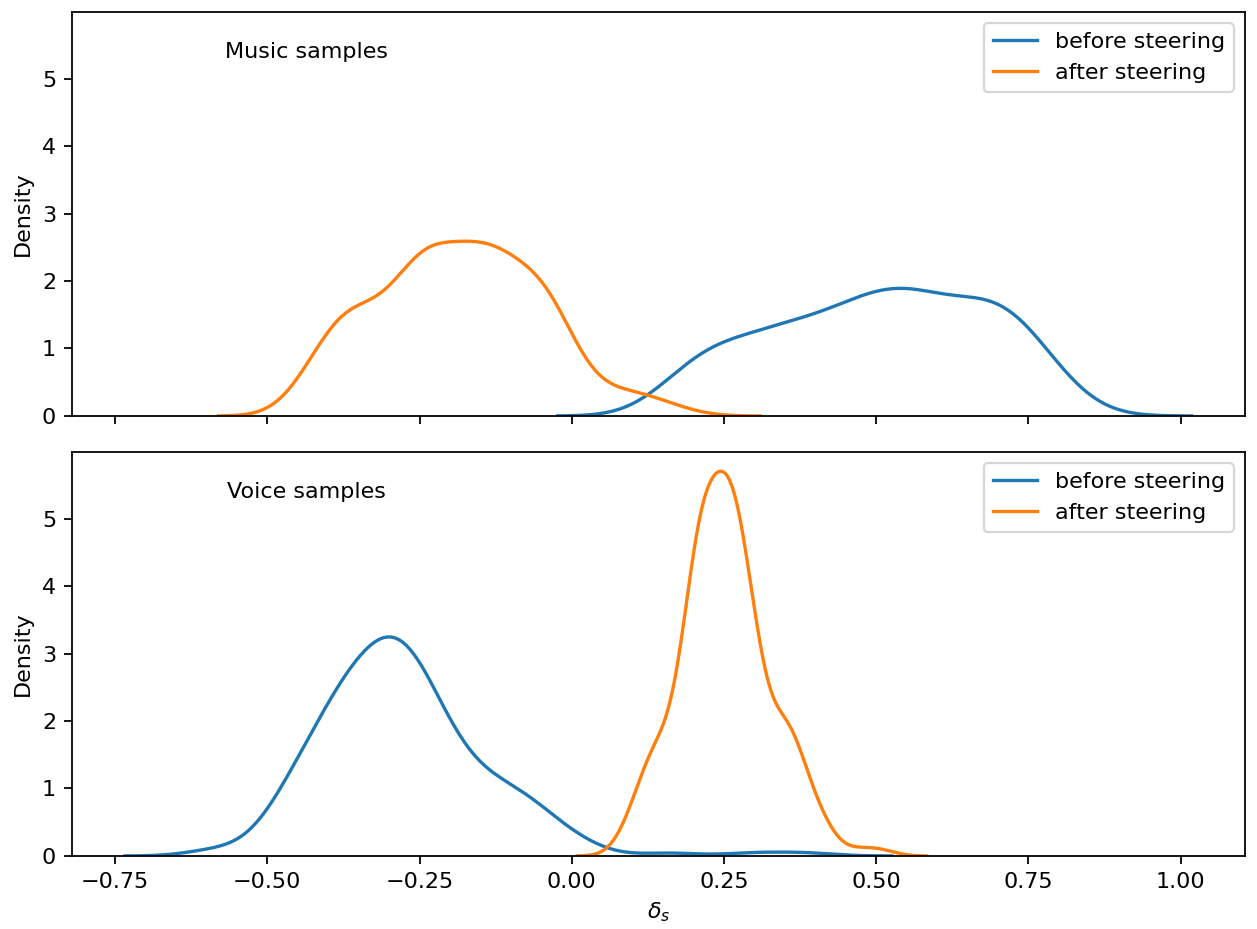}
\caption{Distribution of relative similarity scores before and
after music feature steering.} 
\label{fig:music_steer} 
\end{figure}

Using the findings in Section \ref{sec:id_music} that latent index 74 
represents IVR music in the SAE model with latent dimension of 200 
and $K=20$, feature steering is performed for music and voice samples 
in the Music Feature test set.  The results are presented in Fig.\ref{fig:music_steer} 
and Table \ref{tab:music_steer}, which shows that a music sample embedding
shifts from proximity to the reference music embedding to proximity to
the reference voice embedding when its latent music feature is deactivated,
and a voice sample embedding shifts in the opposite direction when 
its latent music feature is activated.

\begin{table}[htbp]
    \centering
    \begin{tabular}{|c|c|c|}
    \hline
        & Before Steering & After Steering \\ \hline
        Music Samples & 0.505 & -0.185 \\
        Voice Samples & -0.276 & 0.252  \\
        \hline
    \end{tabular}
    \caption{Mean values of the relative similarity scores for music feature steering.}
    \label{tab:music_steer}
\end{table}

\subsection{Discussions} 
\label{sec:discussion}

One substantial difference between this current work and the works on LLM-based
embeddings is the shortage of clearly discrete features. Language is the
primary feature utilized in this work because it is easily discernible in this data and binary. Even the perceived gender of the speaker proved to be
less dualistic than expected, causing difficulty even during human annotation. This labeling challenge is the reason that gender was excluded from the feature identification task.

The remainder of obvious features, such as pitch, volume, prosody, rate of speech,
emotion, etc. exist as a continuum. It is unclear how these continuous features would be
encoded into the latent space. It is possible that binning occurs, but any such
binning is not obvious in the latent spaces investigated here.

Another interesting discussion is on how the data plays a role directly in
which features are created and how they manifest. The majority of the Autoencoder training set is composed of English language audios. It is interesting to note
that there does not seem to be any English feature in any of the autoencoders
that were trained. It would appear that English is more simply encoded as the
default. Spanish therefore is activated to indicate a change from the
default. For example, no French feature is expected, but certainly if French
speakers were present in the dataset used to train the autoencoder, a French
index may appear. In this way, perhaps obviously, the autoencoder reflects not only the
parent model, but also the data on which it is trained.

Anthropic's studies on sparse autoencoders and LLMs lead to several characteristics that were consistent across models that they trained\cite{bricken2022monosemanticity}. These characteristics are summarized as:
\begin{enumerate}
    \item Autoencoders extract monosemantic features
    \item Autoencoders produce interpretable features hidden in the original embeddings
    \item Autoencoder features can be used to steer reconstruction
    \item Features appear to split as latent space grows
    \item Autoencoders produce relatively universal features
    \item A small embedding can represent thousands of latent space features
    \item Features connect in finite-state automata
\end{enumerate}

This present work supports points 1 through 4 for audio data, and we plan to explore 5 through further study.



\subsection{Limitations} 
\label{sec:limitations}
A notable difference in this work compared to the works on LLM-based tokens is the ratio between the dimensions of the latent and the embeddings. In the LLM works, the latent space is several orders of magnitude greater than the input, compared to the largest in this work, which is only approximately 3x greater. Unfortunately, we lack the data to train models of this magnitude.

\section{Conclusion and Future Work}
\label{sec:conclusion}
This study demonstrates the effectiveness of sparse autoencoders in extracting mono-semantic features from non-textual embedded data, specifically speaker embeddings generated from a Titanet model. The results show that the autoencoders exhibit many characteristics found in LLM studies, including feature identification, splitting, and steering. These findings suggest that sparse autoencoders may be a valuable tool for understanding and interpreting embedded data in many domains, including audio-based speaker recognition.


Future work will focus on exploring the universality of the mono-semantic features across different speaker embedding models, as well as investigating the application of sparse autoencoders to Whisper embeddings, which may capture both linguistic and audio characteristics. 


\bibliographystyle{IEEEtran}
\bibliography{sae}

\begin{thebibliography}{10}
\providecommand{\url}[1]{#1}
\csname url@samestyle\endcsname
\providecommand{\newblock}{\relax}
\providecommand{\bibinfo}[2]{#2}
\providecommand{\BIBentrySTDinterwordspacing}{\spaceskip=0pt\relax}
\providecommand{\BIBentryALTinterwordstretchfactor}{4}
\providecommand{\BIBentryALTinterwordspacing}{\spaceskip=\fontdimen2\font plus
\BIBentryALTinterwordstretchfactor\fontdimen3\font minus
  \fontdimen4\font\relax}
\providecommand{\BIBforeignlanguage}[2]{{%
\expandafter\ifx\csname l@#1\endcsname\relax
\typeout{** WARNING: IEEEtran.bst: No hyphenation pattern has been}%
\typeout{** loaded for the language `#1'. Using the pattern for}%
\typeout{** the default language instead.}%
\else
\language=\csname l@#1\endcsname
\fi
#2}}
\providecommand{\BIBdecl}{\relax}
\BIBdecl

\bibitem{Su_2019}
\BIBentryALTinterwordspacing
J.~Su, D.~V. Vargas, and K.~Sakurai, ``One pixel attack for fooling deep neural
  networks,'' \emph{IEEE Transactions on Evolutionary Computation}, vol.~23,
  no.~5, p. 828–841, Oct. 2019. [Online]. Available:
  \url{http://dx.doi.org/10.1109/TEVC.2019.2890858}
\BIBentrySTDinterwordspacing

\bibitem{ribeiro2016whyitrustyou}
\BIBentryALTinterwordspacing
M.~T. Ribeiro, S.~Singh, and C.~Guestrin, ``"why should i trust you?":
  Explaining the predictions of any classifier,'' 2016. [Online]. Available:
  \url{https://arxiv.org/abs/1602.04938}
\BIBentrySTDinterwordspacing

\bibitem{zafar2019dlimedeterministiclocalinterpretable}
\BIBentryALTinterwordspacing
M.~R. Zafar and N.~M. Khan, ``Dlime: A deterministic local interpretable
  model-agnostic explanations approach for computer-aided diagnosis systems,''
  2019. [Online]. Available: \url{https://arxiv.org/abs/1906.10263}
\BIBentrySTDinterwordspacing

\bibitem{shrikumar2019learningimportantfeaturespropagating}
\BIBentryALTinterwordspacing
A.~Shrikumar, P.~Greenside, and A.~Kundaje, ``Learning important features
  through propagating activation differences,'' 2019. [Online]. Available:
  \url{https://arxiv.org/abs/1704.02685}
\BIBentrySTDinterwordspacing

\bibitem{sundararajan2017axiomaticattributiondeepnetworks}
\BIBentryALTinterwordspacing
M.~Sundararajan, A.~Taly, and Q.~Yan, ``Axiomatic attribution for deep
  networks,'' 2017. [Online]. Available: \url{https://arxiv.org/abs/1703.01365}
\BIBentrySTDinterwordspacing

\bibitem{luu2021leveragingspeakerattributeinformation}
\BIBentryALTinterwordspacing
C.~Luu, P.~Bell, and S.~Renals, ``Leveraging speaker attribute information
  using multi task learning for speaker verification and diarization,'' 2021.
  [Online]. Available: \url{https://arxiv.org/abs/2010.14269}
\BIBentrySTDinterwordspacing

\bibitem{wu2024explainableattributebasedspeakerverification}
\BIBentryALTinterwordspacing
X.~Wu, C.~Luu, P.~Bell, and A.~Rajan, ``Explainable attribute-based speaker
  verification,'' 2024. [Online]. Available:
  \url{https://arxiv.org/abs/2405.19796}
\BIBentrySTDinterwordspacing

\bibitem{bricken2022monosemanticity}
T.~Bricken \emph{et~al.}, ``Towards monosemanticity: Decomposing language
  models with dictionary learning,'' \emph{Transformer Circuits Thread}, 2023,
  https://transformer-circuits.pub/2023/monosemantic-features/index.html.

\bibitem{templeton2024scaling}
\BIBentryALTinterwordspacing
A.~Templeton, T.~Conerly, J.~Marcus, J.~Lindsey, T.~Bricken, B.~Chen,
  A.~Pearce, C.~Citro, E.~Ameisen, A.~Jones, H.~Cunningham, N.~L. Turner,
  C.~McDougall, M.~MacDiarmid, C.~D. Freeman, T.~R. Sumers, E.~Rees, J.~Batson,
  A.~Jermyn, S.~Carter, C.~Olah, and T.~Henighan, ``Scaling monosemanticity:
  Extracting interpretable features from claude 3 sonnet,'' \emph{Transformer
  Circuits Thread}, 2024. [Online]. Available:
  \url{https://transformer-circuits.pub/2024/scaling-monosemanticity/index.html}
\BIBentrySTDinterwordspacing

\bibitem{gao2024scalingevaluatingsparseautoencoders}
\BIBentryALTinterwordspacing
L.~Gao \emph{et~al.}, ``Scaling and evaluating sparse autoencoders,'' 2024.
  [Online]. Available: \url{https://arxiv.org/abs/2406.04093}
\BIBentrySTDinterwordspacing

\bibitem{lieberum2024gemmascopeopensparse}
\BIBentryALTinterwordspacing
T.~Lieberum, S.~Rajamanoharan, A.~Conmy, L.~Smith, N.~Sonnerat, V.~Varma,
  J.~Kramár, A.~Dragan, R.~Shah, and N.~Nanda, ``Gemma scope: Open sparse
  autoencoders everywhere all at once on gemma 2,'' 2024. [Online]. Available:
  \url{https://arxiv.org/abs/2408.05147}
\BIBentrySTDinterwordspacing

\bibitem{koluguri2021titanetneuralmodelspeaker}
\BIBentryALTinterwordspacing
N.~R. Koluguri, T.~Park, and B.~Ginsburg, ``Titanet: Neural model for speaker
  representation with 1d depth-wise separable convolutions and global
  context,'' 2021. [Online]. Available: \url{https://arxiv.org/abs/2110.04410}
\BIBentrySTDinterwordspacing

\bibitem{Harper_NeMo_a_toolkit}
\BIBentryALTinterwordspacing
E.~Harper \emph{et~al.}, ``{NeMo: a toolkit for Conversational AI and Large
  Language Models}.'' [Online]. Available: \url{https://github.com/NVIDIA/NeMo}
\BIBentrySTDinterwordspacing

\bibitem{radford2023robust}
A.~Radford, J.~W. Kim, T.~Xu, G.~Brockman, C.~McLeavey, and I.~Sutskever,
  ``Robust speech recognition via large-scale weak supervision,'' in
  \emph{International conference on machine learning}.\hskip 1em plus 0.5em
  minus 0.4em\relax PMLR, 2023, pp. 28\,492--28\,518.

\end{thebibliography}

\end{document}